\documentclass[sigconf]{acmart}
\usepackage[explicit]{titlesec}
\usepackage{multirow}
\usepackage{subfigure}
\usepackage{adjustbox}
\usepackage{float}
\usepackage{array,diagbox,siunitx}
\usepackage{enumitem}
\usepackage{caption}

\usepackage{colortbl}
\DeclareMathOperator*{\argmax}{arg\,max}
\DeclareMathOperator*{\argmin}{arg\,min}
\usepackage{xcolor}
\usepackage[linesnumbered,ruled,vlined]{algorithm2e}

\SetCommentSty{mycommfont}
\SetKwInput{KwInput}{Input}            
\SetKwInput{KwOutput}{Output}
\definecolor{answercolor}{RGB}{240, 240, 240}

\AtBeginDocument{%
  \providecommand\BibTeX{{%
    \normalfont B\kern-0.5em{\scshape i\kern-0.25em b}\kern-0.8em\TeX}}}

\usepackage{colortbl}

\begin{document}

\title[RAL: Sample-Efficient Training]{Robust Active Learning: Sample-Efficient Training of\\Robust Deep Learning Models}

\author{
Yuejun Guo$^{1}$,
Qiang Hu$^{1}$,
Maxime Cordy$^{1}$,
Mike Papadakis$^{1}$,
and Yves Le Traon$^{1}$
\\ 
\normalsize{
$^1$University of Luxembourg, Luxembourg
}
}
\settopmatter{printacmref=false}

\begin{abstract}
Active learning is an established technique to reduce the labeling cost to build high-quality machine learning models. A core component of active learning is the acquisition function that determines which data should be selected to annotate. State-of-the-art acquisition functions -- and more largely, active learning techniques -- have been designed to maximize the clean performance (e.g. accuracy) and have disregarded robustness, an important quality property that has received increasing attention. Active learning, therefore, produces models that are accurate but not robust.

In this paper, we propose \emph{robust active learning}, an active learning process that integrates adversarial training -- the most established method to produce robust models. Via an empirical study on 11 acquisition functions, 4 datasets, 6 DNN architectures, and 15105 trained DNNs, we show that robust active learning can produce models with the robustness (accuracy on adversarial examples) ranging from 2.35\% to 63.85\%, whereas standard active learning systematically achieves negligible robustness (less than 0.20\%). Our study also reveals, however, that the acquisition functions that perform well on accuracy are worse than random sampling when it comes to robustness. We, therefore, examine the reasons behind this and devise a new acquisition function that targets both clean performance and robustness. Our acquisition function -- named density-based robust sampling with entropy (DRE) -- outperforms the other acquisition functions (including random) in terms of robustness by up to 24.40\% (3.84\% than random particularly), while remaining competitive on accuracy. Additionally, we prove that DRE is applicable as a test selection metric for model retraining and stands out from all compared functions by up to 8.21\% robustness.

\end{abstract}

\maketitle

\section{Introduction}
\label{sec:intro}
Deep learning (DL) systems are increasingly present in modern software \cite{Arpteg2018SE,devanbu2020deep}. Aware of the great potential of these systems, big companies such as Google, Microsoft, and Facebook take huge engineering efforts to build and maintain DL-based tools and contribute to the development of the DL field. DL techniques are also useful to support software engineering tasks, e.g., natural language processing for source code understanding \cite{alon2019code2vec}, image recognition for software security \cite{yu2021layout,alahmadi2020code}, and image classification for bugs detection in mobile apps \cite{Wang2021robot}.

However, one of the most important engineering hurdles in developing a DL model (i.e., a Deep Neural Network (DNN)) with competitive performance is the large amount of labeled data required for training. Although massive data can come at little to no cost, labeling all of them is time-consuming and almost impossible, especially when expert knowledge is indispensable. To overcome this problem, research has proposed approaches to maximize learning capabilities in situations where unlabeled data are abundant while labeling capacities are limited. 

Active learning is such an approach and has been widely applied in various domains, such as image classification \cite{malanie2018dfal}, information extraction \cite{settles2008active}, speech recognition \cite{zhu2005semi}, software defect prediction \cite{Lu2014defect}, and program behavior classification \cite{bowring2004behavior}. Active learning aims at training a DL model using a limited number of but carefully selected labeled data, such that this model can achieve a similar performance compared to the model that would be trained on all (labeled) data. This process is usually iterative: at each iteration (stage), a so-called \emph{acquisition function} determines the set of unlabeled data to use and an oracle (for instance, a human annotator) is queried to provide the labels of these data. Past studies have shown that active learning has the potential to reach the same performance as a model trained with all data labeled using only 20\% of these data (for the SVHN dataset) \cite{sener2018active}.

One important limitation of active learning so far is that it focuses only on \emph{clean performance} (most often, test accuracy \cite{malanie2018dfal,burr2008EGL,sener2018active}) and ignores other important quality indicators. In particular, active learning does not produce models that are robust to \emph{adversarial examples} -- a fact that we empirically confirm in this paper. These examples are produced by introducing a small perturbation to benign inputs in a way that it changes the decision of the model although it should not \cite{Szegedy2014intriguing}. Recent researches revealed that DNNs are highly vulnerable to adversarial examples created using automated procedures \cite{Szegedy2014intriguing,Goodfellow2015adv,Schmidt2018adv,hu2020adversarial}. Adversarial examples, therefore, raise major security concerns \cite{bojarski2016end,yuan2014droid} and have gained increasing attention. It is challenging to address that successive methods improve models' robustness (accuracy on adversarial examples) merely by a small percent \cite{croce2020robustbench}.

In this paper, we bootstrap an endeavor towards improving the adversarial robustness of DNNs trained through active learning. We propose a new training process -- named \emph{robust active learning} -- which merges active learning with adversarial training \cite{madry2018towards}, the most effective defense against adversarial examples. The key idea of robust active learning is to iteratively select the data to label (just like in ``standard active learning'') and, then, to train the model not on the original data but on their adversarial counterparts. 

An essential question that we need to answer is whether existing acquisition functions remain effective in robust active learning. We, therefore, conduct the first comprehensive study that evaluates whether these acquisition functions can produce models that are both accurate \emph{and} robust. We carry out experiments on four datasets, six DNN architectures, eleven acquisition functions, and dozens of labeling budgets. Our investigations reveal that while state-of-the-art acquisition functions remain effective when it comes to clean performance, when it comes to robustness all of them are outperformed by random sampling.

Following our findings, we explore what key factors can explain the lack of robustness of models trained using existing acquisition functions. We empirically demonstrate that, while acquisition functions are inherently biased towards selecting data with specific characteristics (e.g. data with the highest entropy), this bias strongly negatively correlates with robustness. This implies that, in robust active learning, effective acquisition functions should select a subset of data that are representative (with respect to the aforementioned characteristics) of the whole dataset. 

As a result, we propose a new acquisition function -- named \emph{density-based robust sampling with entropy (DRE)} -- that selects data while minimizing the difference between the entropy distributions of the selected set and full dataset. We compare DRE with all existing acquisition functions. Our results demonstrate that DRE is the sole acquisition function that achieves higher robustness than random while being competitive in terms of accuracy.

While robust active learning aims at training robust models from scratch, an adjacent problem is DL testing, i.e. the problem of finding (adversarial) examples that a trained (high accuracy) DL model misclassifies. Such examples can then be used to retrain the model, thereby improving its adversarial robustness. A crucial part of DL testing is the selection of data from which to generate the adversarial examples. The same acquisition functions that are used in active learning can also serve this purpose. We, therefore, investigate the effectiveness of the acquisition functions in driving the DL testing and retraining. Our evaluation demonstrates that DRE achieves again the highest gains in robustness.

To sum up, the main contributions of this paper are:
\begin{enumerate}[nosep]
    \item We are the first to formulate and investigate robust active learning, i.e. the problem of training accurate and robust models with a limited budget of labeled data.
    \item We conduct the first empirical study on the effectiveness of existing acquisition functions in terms of accuracy and robustness. Our results demonstrate that, though some acquisition functions yield higher accuracy than random sampling (by up to 8.08\%), none of the functions we investigate outperforms random sampling in terms of robustness (by up to 22.50\%).
    \item We demonstrate that the inherent bias of acquisition functions towards the ``informative data'' is the cause for their lesser robustness (compared to random).
    \item We propose DRE, a new acquisition function that yields better robustness (by up to 24.40\%) than existing functions -- 3.84\% than the best (random) -- while achieving competitive accuracy. DRE, therefore, constitutes the best compromise between accuracy and robustness and forms a new baseline for future research on robust active learning.
    \item We investigate the use of DRE in the adjacent problem of test selection for model retraining. We show that DRE achieves competitive accuracy and outperforms other acquisition functions by up to 8.21\% robustness.
\end{enumerate}

\section{Background and Related work}
Here, we introduce the relevant related work on active learning, adversarial robustness, and DL testing. Readers can refer to several surveys \cite{ren2020survey,Kui2020survey,settles2009al,tong2001active,Burr2010survey,Zhang2019survey} for comprehensive surveys of these topics.

\subsection{Active Learning}
\label{subsec:activeLearning}
Active learning is a set of techniques that aims at building high-performing models using only a small set of labeled data. The main hypothesis is that if the model learns from the most informative data, it can obtain a similar performance using substantially fewer data compared with learning on the entire training set.

\paragraph{\textbf{Problem scenario and active learning in deep learning}} Typically, in active learning, there are three types of problem scenario, membership query synthesis, selective sampling, and pool-based sampling \cite{settles2009al,Kui2020survey}. In the case of membership query synthesis, the model generates data to query instead of choosing data from the available unlabeled set \cite{angluin1988queries}. Selective sampling also refers to stream-based or sequential active learning \cite{atlas1990training}. In this scenario, the model receives unlabeled data once at a time and decides whether or not to request the label based on the informativeness. Compared with selective sampling, in pool-based sampling, the model requests labels of a collection of unlabeled data at once in each query \cite{lewis1994sequential}. Since in deep learning, data are divided into batches to train DNNs and a single data would not impart significant change in the model \cite{Szegedy2014intriguing}, the pool-based sampling is the most widely applied active learning scenario. We, therefore, focus on pool-based sampling.

\paragraph{\textbf{Empirical study}} As many approaches (acquisition functions) exist for active learning, several empirical studies have been conducted to compare their effectiveness. Arguing that most previous empirical studies were limited to a single model and performance metric, Ramirez-Loaiza \emph{et al.} \cite{ramirez2017active} rather employ two models and five measures in their study. However, all the used metrics (i.e. precision, recall, F1, accuracy, and AUC) measures the correctness of the model on clean data only. \cite{pereira2019empirical} studies more datasets, classification models, and selection approaches, but the comparison is still based on correctness only. On the other hand, some studies focus on specific applications, such as segmenting Japanese word segmentation \cite{sassano2002empirical}, text classification \cite{Prabhu2019bias}, learning English verb senses \cite{chen2006empirical}, labeling sequence \cite{settles2008analysis}, and locating temporal activation in video data \cite{heilbron2018annotate}.

In this paper, we overcome the two main limitations of existing studies. First, we go beyond prediction correctness and consider adversarial robustness as an important success metric of active learning. Second, we compare 11 competitive active learning approaches on common ground.

\paragraph{\textbf{Active learning for SE}} Many tasks in software engineering (SE) benefit from active learning. For instance, Bowring \emph{et al.} \cite{bowring2004behavior} applies active learning in the automatic classification of program behavior. The model is incrementally trained using selected executions that represent unknown behaviors for the model. When performing on the software effort estimation data, active learning is proved to largely prune the training data and quickly find the essential content \cite{Ekrem2013essential}. Yu \emph{et al.} \cite{yu2018finding} find that active learning can help with conducting literature reviews. Recently, Cambronero \emph{et al.} \cite{Cambronero2019active} propose to use active learning to automatically infer programs, which can significantly alleviate the difficulty of manual annotations. Based on the Mozilla Firefox vulnerability data, Yu \emph{et al.} \cite{Yu2019improving} prove that active learning is useful in building a prediction model which learns from the historical source code data and predicts the candidates to inspect. Similarly, Yang \emph{et al.} \cite{yang2021static} study active learning for static code analysis. Active learning also facilitates building defect prediction models \cite{lu2012fault,lu2015thesis,Lu2014defect,Tu2020better}. For instance, Tu \emph{et al.} \cite{Tu2020better} develop an active learning tool, EMBLEM, to label the most problematic commits and they claim the first use of active learning in commit defect prediction.

\subsection{Adversarial Robustness}
\label{subsec:advRobust}
The adversarial robustness of DNNs relates to the ability of the model to distinguish adversarial examples. In other words, similar to the accuracy on clean data, the robustness of a DNN is measured by its accuracy on adversarial examples crafted from the clean data. 

\paragraph{\textbf{Adversarial example}} Taking the image classification task as an example, given an input image, the corresponding adversarial example is crafted by adding a carefully calculated perturbation into this image to mislead DNNs \cite{Szegedy2014intriguing}. Since this perturbation is hardly perceptible for human beings, an adversarial example is regarded as following the same data distribution and having the same label as the original input. The cause of adversarial examples is still under exploration. Some speculative explanations are the extreme non-linearity of DNNs \cite{Szegedy2014intriguing}, the high-dimensional linearity of DNNs \cite{Goodfellow2015adv}, the presence of non-robust features \cite{hu2020adversarial}, insufficient regularisation \cite{Schmidt2018adv}, and insufficient model averaging \cite{Goodfellow2015adv}. 

\paragraph{\textbf{Adversarial attack}} The approaches to craft adversarial examples are called adversarial attacks (also refer to adversaries, threat models). In general, adversarial attacks are divided into three types: black-box, gray-box, and white-box. The black-box attacks have no access to the model, and the perturbation is calculated by using the predicted probabilities or logits (score-based), e.g., square attack \cite{andriushchenko2020square}, by simply relying on the max class label (decision-based) \cite{BrendelRB18boundary}, or by transferring from another DNN model trained with the same training data (transfer-based) \cite{Bhagoji2018ECCV}. The gray-box attacks only have knowledge of DNN architectures. In contrast, the white-box attacks have full access to the data, parameters, and DNN architectures. The gradient of training loss is mostly utilized in various white-box methods, e.g., projected gradient descent (PGD) \cite{madry2018towards}. Besides, there is the Auto attack \cite{croce2020reliable}, which combines the black-box and white-box attacks and is commonly used as a strong baseline in robustness evaluation. In this paper, we comprehensively consider the square attack (black-box), PGD (white-box) attack, and Auto attack (adaptive).

\paragraph{\textbf{Adversarial defense}} Mitigating the threat of adversarial examples and improving the robustness of DNNs is of great importance and has attracted considerable attention \cite{Kui2020survey}. Adversarial defense refers to a defense mechanism that secures DNNs against adversarial attacks. Many defense approaches have been proposed, such as adversarial training \cite{madry2018towards}, input denoising \cite{defensegan}, input transformation \cite{guo2018countering}, randomization \cite{xie2018mitigating}, and defensive distillation \cite{Distillation2016,WardeFarley20161AP}. Among all, adversarial training with a PGD adversary \cite{madry2018towards} has been proven to be one of the most effective methods. Compared with standard training where the original training data is fed into the DNN models to tune all the parameters, in each epoch of adversarial training, a new set of data is generated by applying the PGD attack to the original training data and then is used to train the model.  

\subsection{Test Selection for Retraining}
\label{subsec:testSelection}
Software engineering researchers have devoted substantial effort to evaluate and improve the quality of DL models. DL testing techniques aim to expose flaws in DL models through either the selection of test data (from an existing data pool) or the generation of new test data (i.e. adversarial examples). DL engineers can, then, use those data to retrain the model and improve it. 

Multiple test selection metrics have been proposed to solve various DL testing problem, including test accuracy estimation \cite{10.1145/3394112, 10.1145/3338906.3338930}, test data prioritization \cite{prima2021,yang2020gini,ma2021testselection}, DNN comparison \cite{meng2021compare}, and DNN retraining \cite{Shen2020mcp,Wang2021robot, yang2020gini}.

In particular, past researches on DNN retraining \cite{yang2020gini, Shen2020mcp} have shown that, given a trained DNN and a set of unlabeled test data, test selection metrics can select data to retrain the DNN (either on the selected data or on adversarial examples produced from the data) and improve its quality (clean performance or robustness). This testing and retraining process shares the similarity with robust active learning as we define it in this paper. Indeed, both processes aim to improve the quality of DNNs through the selection of the most appropriate data. The difference is that active learning applies during training and targets the training data, whereas DL testing applies after training and on a set of data not used during training. Therefore, the retraining process can be seen as one-stage active learning after training and on a new set of data. The acquisition functions used in active learning -- including the one that we propose in this paper -- can thus be used for retraining as well.

\section{Motivation and Problem Formulation}
\label{sec:pro}
Consider a $C$-class classification problem over a sample space $\mathcal{Z}=\mathcal{X}\times\mathcal{Y}\rightarrow\mathcal{R}$, and a collection of data $\mathcal{D}=\left\{x_i,y_i\right\}_{i\in[n]}\sim p_{\mathcal{Z}}$, where $x_i\in\mathcal{X}$ is the data and $y_i\in\mathcal{Y}$ is the corresponding label. Let $\mathcal{H}$ be the hypothesis space and $f_\theta$ be a neural network architecture parameterized by a parameter vector $\theta\in\mathcal{H}$. We define the loss function as $J:\mathcal{H}\times\mathcal{Z}\rightarrow\mathcal{R}$. In deep learning, the object of training $f_\theta$ is to minimize the expected loss (also called as the expected risk):
\begin{equation}
    \mathbb{E}_{x,y\sim p_{\mathcal{Z}}}\left[J\left(\theta_{\mathcal{D}},x,y\right)\right]
\end{equation}
where $\theta_{\mathcal{D}}$ indicates that the parameters are tuned using $\mathcal{D}$. Note that the value of $\theta$ depends on the data used to train, thus, to put this in evidence, we use the notation $\theta_{\mathcal{D}}$ indicating that this parameter vector corresponds to the data $\mathcal{D}$. 

\begin{figure}
    \centering
    \includegraphics[scale=0.44]{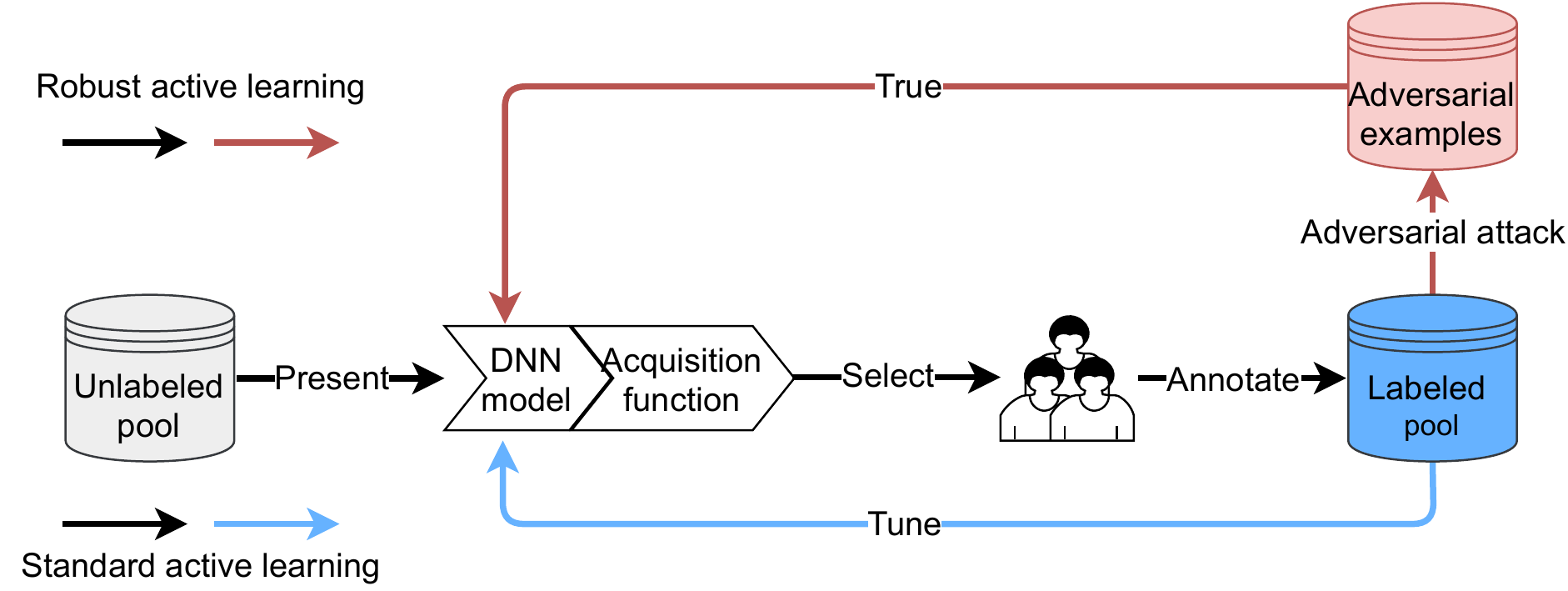}
    \caption{Overview of pool-based active learning.}
    \label{fig:al}
\end{figure}

Figure \ref{fig:al} illustrates our proposed process of robust active learning and its difference to standard active learning. The data are initially unlabeled and stored in an unlabeled pool, $\mathcal{U}=\left\{x_i\right\}_{i\in\left[n\right]}\sim p_{\mathcal{Z}}$. Given a labeling budget $b$, an acquisition function calculates the priority of each data to be labeled and the process requests some oracle to label the highest priority data. These newly labeled data are then moved to a labeled pool $\mathcal{L}=\left\{x_i,y_i\right\}_{i\in\left[m\right]}\sim p_{\mathcal{Z}}$ ($m\leq b$). In standard active learning, at each stage, the set of model parameters $\theta$ is updated using all data from the labeled pool -- so as to minimize the average model loss on these labeled data, that is,
\begin{equation}
\label{equ_al}
    \mathbb{E}_{x,y\sim p_{\mathcal{Z}}}\left[J\left(\theta_{\mathcal{L}},x,y\right)\right]
\end{equation}

Just like DNNs trained on clean data are well-known to be vulnerable to adversarial examples \cite{Szegedy2014intriguing,Goodfellow2015adv,hu2020adversarial,Schmidt2018adv}, DNNs trained with standard active learning should have low robustness as well. To confirm this, we conducted experiments on six subjects to measure the robustness of models that result from standard active learning using different acquisition functions (following the experimental protocol described in Section \ref{sec:exp}). Table \ref{tab:exp1-robu} summarizes the results. We observe that in all cases, the achieved robustness is extremely low (below 0.20\%), while the highest accuracy approaches 100\%.

\begin{table*}
\caption{Standard active learning: result of accuracy (\%) and adversarial robustness (\%) against the Auto attack by 11 acquisition functions. Maximum: the maximum in a dataset.}
\label{tab:exp1-robu}
\resizebox{\linewidth}{!}{
\begin{tabular}{|l|rr|rr|rr|rr|rr|rr|}
\hline
\multirow{2}{*}{\textbf{\begin{tabular}[c]{@{}l@{}}Acquisition\\ Function\end{tabular}}} & \multicolumn{2}{c|}{\textbf{MNIST: Lenet-5}} & \multicolumn{2}{c|}{\textbf{Fashion-MNIST: Lenet-5}} & \multicolumn{2}{c|}{\textbf{SVHN: VGG8}} & \multicolumn{2}{c|}{\textbf{CIFAR-10: VGG16}} & \multicolumn{2}{c|}{\textbf{CIFAR-10: ResNet18}} & \multicolumn{2}{c|}{\textbf{CIFAR-10: PreActResNet18}} \\ \cline{2-13} 
 & \textbf{Accuracy} & \textbf{Robustness} & \textbf{Accuracy} & \textbf{Robustness} & \textbf{Accuracy} & \textbf{Robustness} & \textbf{Accuracy} & \textbf{Robustness} & \textbf{Accuracy} & \textbf{Robustness} & \textbf{Accuracy} & \textbf{Robustness} \\ \hline
BALD & 99.61 & 0.00 & 87.65 & 0.00 & 90.85 & 0.08 & 93.04 & 0.11 & 94.81 & 0.16 & 94.29 & 0.19 \\
DFAL & 99.45 & 0.00 & 87.73 & 0.00 & 90.61 & 0.07 & 93.25 & 0.08 & 94.36 & 0.20 & 94.38 & 0.09 \\
EGL & 99.61 & 0.00 & 84.73 & 0.00 & 88.69 & 0.05 & 92.65 & 0.08 & 94.19 & 0.15 & 94.40 & 0.11 \\
MaxEntropy & 99.69 & 0.00 & 87.57 & 0.00 & 91.05 & 0.05 & 93.21 & 0.08 & 94.53 & 0.11 & 94.63 & 0.09 \\
DropOut-Entropy & 99.59 & 0.00 & 87.28 & 0.00 & 91.03 & 0.05 & 93.27 & 0.07 & 94.60 & 0.11 & 94.45 & 0.13 \\
DeepGini & 99.63 & 0.00 & 87.87 & 0.00 & 90.77 & 0.04 & 93.45 & 0.11 & 94.51 & 0.19 & 94.65 & 0.14 \\
Core-set & 99.67 & 0.00 & 87.64 & 0.00 & 90.81 & 0.06 & 93.17 & 0.08 & 94.29 & 0.17 & 94.55 & 0.20 \\
LC & 99.59 & 0.00 & 87.98 & 0.00 & 90.91 & 0.06 & 93.20 & 0.16 & 94.57 & 0.17 & 94.47 & 0.12 \\
Margin & 99.63 & 0.00 & 87.86 & 0.00 & 90.95 & 0.08 & 93.09 & 0.10 & 94.51 & 0.15 & 94.53 & 0.17 \\
MCP & 99.54 & 0.00 & 87.55 & 0.00 & 90.45 & 0.07 & 92.76 & 0.08 & 94.33 & 0.16 & 94.05 & 0.14 \\
Random & 98.84 & 0.00 & 85.73 & 0.00 & 87.53 & 0.11 & 91.09 & 0.16 & 92.23 & 0.19 & 92.27 & 0.14 \\ \hline
\textbf{Maximum} & \textbf{99.69} & \textbf{0.00} & \textbf{87.98} & \textbf{0.00} & \textbf{91.05} & \textbf{0.08} & \textbf{93.45} & \textbf{0.20} & \textbf{94.81} & \textbf{0.20} & \textbf{94.65} & \textbf{0.20} \\ \hline
\end{tabular}}
\end{table*}

To facilitate the trained DNN to be resilient to adversarial examples, we propose to incorporate the adversarial training into active learning. Compared with standard active learning, for each labeled data we craft an adversarial example -- using the PGD attack \cite{madry2018towards} -- and train the DNN with the produced examples (instead of the original labeled data). Hence, the clean labeled data $\{(x_i, y_i)\}$ are replaced by perturbed examples $\mathcal{L}_{adv}=\left\{x_i+\delta,y_i\right\}_{i\in\left[m\right]}$ ($m\leq b$) during training and the training objective becomes to minimize the risk concerning the adversarial examples:
\begin{equation}
\label{equ_rac}
    \mathbb{E}_{x,y\sim p_{\mathcal{Z}}}\left[J\left(\theta_{\mathcal{L}_{adv}},x+\delta,y\right)\right]
\end{equation}
where $\delta$ is the upper bound of the perturbation. Remark that in adversarial training, the perturbed data produced are used even if the DNN correctly classifies them (i.e. even if the adversarial attack algorithm ``fails'').

\vspace{2mm}
\noindent\fcolorbox{black}{answercolor}{\begin{minipage}{0.97\columnwidth}
The \textbf{central research question} that we study in this paper is whether robust active learning can produce robust models even though it does not make use of all data available. Of course, we expect that the answer to this question largely depends on the used acquisition function. We, therefore, conduct a comprehensive empirical study involving a large set of acquisition functions.
\end{minipage}}

\section{Experimental Protocol}
\label{sec:exp}
\subsection{Implementation and Hardware}
All experiments were conducted on a high-performance computer cluster and each cluster node runs a 2.6 GHz Intel Xeon Gold 6132 CPU with an NVIDIA Tesla V100 16G SXM2 GPU. We implement the proposed approach and existing acquisition functions based on the state-of-the-art framework, PyTorch 1.6.0. To allow for reproducibility, our full implementation and evaluation subjects are available on GitHub \footnote{We will make the implementation publicly available upon acceptance.}. To reduce the influence of randomness, we repeat each experiment three times and, in total, 15105 DNNs are involved in this empirical study. Due to the space limitation, we only report the results when model robustness is evaluated against the Auto attack, the strongest attack that we considered. Also, in Section 6 our investigations are illustrated for all datasets only. The remaining results corroborate our findings and are freely available on our companion project website \cite{homepage}. 

\subsection{Datasets and Models}
We select four popular publicly-available datasets, MNIST \cite{Lecun1998gradient}, Fashion-MNIST \cite{fashionMnist}, SVHN \cite{Netzer2011reading}, and CIFAR-10 \cite{Alex2009techm} for evaluation. These datasets have previously been used for robustness assessment in the context of DL testing \cite{Wang2021robot}. Both MNIST and Fashion-MNIST include a collection of 28 x 28 grayscale images. MNIST comprises handwritten digits $0\sim 9$ and Fashion-MNIST presents fashion products. SVHN and CIFAR-10 consist of 32 x 32 RGB images. SVHN shows street view house numbers and CIFAR-10 contains more complex entities, such as vehicles and animals. Table \ref{tab:datasets} summarizes the detailed information of the datasets and DNN models. In all our experiments, for each dataset, we use the training set to craft the adversarial examples during the adversarial training. We evenly split the test data into a validation set and a test set. The validation set is used in the training process to tune the parameters of DNNs, and the test set is independent of training for evaluating the accuracy (directly on the test data) and the robustness (by crafting adversarial examples from the test data).

\begin{table}
    \caption{Summary of datasets and DNNs. \textbf{Accuracy}: accuracy (\%) of DNNs by standard training using the entire set.}
    \label{tab:datasets}
    \resizebox{\columnwidth}{!}{
    \begin{tabular}{|l|l|r|r|r|}
\hline
\textbf{Dataset} & \textbf{DNN} & \textbf{Train size} & \textbf{Test size} & \textbf{Accuracy} \\ \hline
MNIST & Lenet-5 & 60k & 10k & 99.47 \\
Fashion-MNIST & Lenet-5 & 60k & 10k & 90.78 \\
SVHN & VGG8 & 50k & 10k & 92.69 \\
CIFAR-10 & VGG16 & 50k & 10k & 89.75 \\
 & ResNet18 & 50k & 10k & 90.65 \\
 & PreActResNet18 & 50k & 10k & 90.68 \\ \hline
\end{tabular}
}
\end{table}

\subsection{Robust Active Learning Process and Test Selection}
Table \ref{tab:al} summarizes the active learning setups. Concretely, we follow \cite{Mayer2020asal} to initialize the labeled pool with a small amount of data uniformly sampled from the unlabeled pool. For all the acquisition functions, the initial DNN is the same tuned using these initial sets via standard training. 

\begin{table*}
\caption{Configuration setting of robust active learning. Budget: maximum number of data to label; Initial: number of labeled data when launching active learning; New: number of data to annotate in each stage; Stage: number of stages in active learning; Initial accuracy (\%): accuracy of the initial model; Initial robustness (\%): robustness of initial model evaluated by the PGD attack, square attack, and Auto Attack, respectively. Full: DNN is adversarially trained using the entire data.}
\label{tab:al}
\resizebox{1\textwidth}{!}{
\begin{tabular}{|l|l|r|r|r|r|r|rrr|r|rrr|}
\hline
\multirow{2}{*}{\textbf{Dataset}} & \multirow{2}{*}{\textbf{DNN}} & \multirow{2}{*}{\textbf{Budget}} & \multirow{2}{*}{\textbf{Initial}} & \multirow{2}{*}{\textbf{New}} & \multirow{2}{*}{\textbf{Stage}} & \multirow{2}{*}{\textbf{Initial accuracy}} & \multicolumn{3}{r|}{\textbf{Initial robustness}} & \multirow{2}{*}{\textbf{Full accuracy}} & \multicolumn{3}{r|}{\textbf{Full robustness}} \\ \cline{8-10} \cline{12-14} 
 &  &  &  &  &  &  & \textbf{PGD} & \textbf{Square} & \textbf{Auto} &  & \textbf{PGD} & \textbf{Square} & \textbf{Auto} \\ \hline
MNIST & Lenet-5 & 5k & 200 & 200 & 24 & 29.65 & 0.00 & 0.04 & 0.00 & 98.69 & 89.52 & 76.27 & 76.17 \\
Fashion-MNIST & Lenet-5 & 6k & 200 & 200 & 29 & 74.22 & 0.00 & 0.04 & 0.00 & 72.55 & 64.18 & 34.01 & 34.06 \\
SVHN & VGG8 & 10k & 1k & 500 & 18 & 51.21 & 0.34 & 1.92 & 0.25 & 83.95 & 40.77 & 37.50 & 35.94 \\
CIFAR-10 & VGG16 & 25k & 1k & 500 & 48 & 37.01 & 1.71 & 4.27 & 1.54 & 69.54 & 42.52 & 43.41 & 40.71 \\
 & ResNet18 & 25k & 1k & 500 & 48 & 31.69 & 1.40 & 3.73 & 1.21 & 73.11 & 42.52 & 43.41 & 42.65 \\
 & PreActResNet18 & 25k & 1k & 500 & 48 & 33.88 & 2.51 & 4.76 & 2.23 & 73.24 & 44.29 & 45.59 & 42.65 \\ \hline
\end{tabular}
}
\end{table*}

Table \ref{tab:attacks} lists the parameters for robust active learning. We employ both white-box and black-box attacks for robustness evaluation. The white-box attack PGD is commonly used in the evaluation of DNNs' robustness. Considering that in this case, the defense DNN knows this attack in advance, we also consider the state-of-the-art black-box attack named square attack \cite{andriushchenko2020square}. Finally, we use the Auto attack \cite{croce2020reliable}, an adaptive attack that is widely utilized as a strong baseline thanks to its ability to overcome common defense mechanisms such as gradient masking \cite{athalye2018Obfuscated}.  
All these attacks are implemented using a public PyTorch library, Torchattacks \cite{kim2020torchattacks}. We use the default setting in the library for the other related parameters in each attack.

\begin{table}
\caption{Configurations of adversarial training and evaluation. $\epsilon$, $\alpha$, and $I$ denote the perturbation size, step size of perturbation, and the number of iterative steps, respectively.}
\label{tab:attacks}
\resizebox{\columnwidth}{!}{
\begin{tabular}{|l|l|r|r|}
\hline
\textbf{Operation} & \textbf{Attack} & \textbf{\begin{tabular}[c]{@{}r@{}}MNIST\\ Fashion-MNIST\end{tabular}} & \textbf{\begin{tabular}[c]{@{}r@{}}SVHN\\ CIFAR-10\end{tabular}} \\ \hline
\begin{tabular}[c]{@{}l@{}}Adversarial\\ Training\end{tabular} & PGD & $\epsilon$=0.3, $\alpha$=0.01, $I$=40 & $\epsilon$=0.3, $\alpha$=0.01, $I$=40 \\ \hline
\multirow{3}{*}{\begin{tabular}[c]{@{}l@{}}Robustness\\ Evaluation\end{tabular}} & PGD & $\epsilon$=0.3, $\alpha$=0.01, $I$=50 & $\epsilon=8/255$, $\alpha$=2/255, $I$=50 \\
 & Square & $\epsilon$=0.3 & $\epsilon$=8/255 \\
 & Auto & $\epsilon$=0.3 & $\epsilon$=8/255 \\ \hline
\end{tabular}
}
\end{table}

In test selection for model retraining, we train for 20 epochs for SVHN and 10 for the others to ensure the training process converges.

\subsection{Acquisition Functions}
\label{sec:functions}
Given a DNN $f_\theta$ where $\theta$ is randomly initialized, unlabeled pool $\mathcal{U}$, an acquisition function helps the AL system in each step to query the most informative and useful data to solve the optimization object in Equation \ref{equ_al}. Various acquisition functions have been proposed for data selection. In this section, we review 11 widely used acquisition functions on which we conduct our experiments. 

As Wald showed \cite{Wald1945statis}, a solution to solve the optimization object in Equation \ref{equ_al} is to minimize the maximum of the risk. Based on this, several acquisition functions have been proposed to select the most informative data by assigning and ranking the importance of data. We denote $\left\{p\left(y=i\mid x;\theta\right)\right\}_{i\in[C]}$ be the softmax output of $f_\theta$ (recall that $C$ is the number of classes).

\textbf{1) MaxEntropy}. In information theory, the entropy (also known as Shannon entropy) quantifies the uncertainty of prediction \cite{Shannon1948entropy}:

\begin{equation}
    \label{equ_entropy}
    H\left(y\mid x;\theta\right)=-\sum_{i\in\left[C\right]}p\left(y=i\mid x;\theta\right)\log\left(p\left(y=i\mid x;\theta\right)\right)
\end{equation}
MaxEntropy ranks the uncertainty of data based on the predictive entropy and selects the most uncertain ones.

\textbf{2) DeepGini$^*$}. In decision tree learning, the Gini impurity is a loss metric to decide the optimal split from a root node and subsequent splits. It measures the likelihood of misclassification of a new instance, which in other words reflects the uncertainty of a model to this instance. Borrowing the idea of Gini impurity to the deep neural network, DeepGini \cite{yang2020gini} is proposed to select the most informative data:
\begin{equation}
    \label{equ_gini}
    \underset{{\bf{x}}\in{\mathcal{U}}}{\argmax}\left(1-\sum_{i\in\left[C\right]}p^{2}\left(y=i\mid x;\theta\right)\right)
\end{equation}
Similar to MaxEntropy, DeepGini utilizes the output of $f_\theta$ to assign the informativeness to data. However, as mentioned by the authors of DeepGini, computing the quadratic sum is easier and simpler than performing the logarithmic computation, which is supposed to give better performance.

\textbf{3) BALD} Taking the concept of Bayesian neural network where a DNN is defined by a set of parameters drawn from a posterior distribution, the Bayesian active learning by disagreement (BALD) \cite{houlsby2011bayesian} seeks data for which the parameters under the posterior disagree about the prediction the most. 
\begin{equation}
    \label{equ_bald}
    \underset{{\bf{x}}\in{\mathcal{U}}}{\argmax}\left(H\left(y\mid x;\mathcal{U}\right)-\mathbb{E}_{\theta\sim p_{\theta,\mathcal{U}}}\left[H\left(y\mid x;\theta\right)\right]\right)
\end{equation}
In other words, BALD can be explained as identifying on which the DNN is on average most uncertain about the prediction (big $H\left(y\mid x;\mathcal{U}\right)$) but existing model parameters are confident (small $\mathbb{E}_{\theta\sim p_{\theta,\mathcal{U}}}\left[H\left(y\mid x;\theta\right)\right]$). In practice, to approximate the inferences, the Monte Carlo dropout is widely applied due to its great capacity and low cost. We set $T=10$ with probability 0.1 at test time to sample different DNNs as \cite{malanie2018dfal}.

\textbf{4) DropOut-Entropy} Instead of computing the entropy over one DNN with fixed parameters, this acquisition \cite{Gal2017drop} function calculates the uncertainty over multiple Bayesian DNNs inferred by the Monte Carlo dropout. Thus, the object changes to find data that maximize the uncertainty as follows:
\begin{equation}
    \label{equ_drop}
    H\left(y\mid x;\mathcal{H}\right)=-\sum_{i\in\left[C\right]}p\left(y=i\mid x;\mathcal{H}\right)\log p\left(y=i\mid x;\mathcal{H}\right)
\end{equation}
where $p\left(y=i\mid x;\mathcal{H}\right)=\frac{1}{T}\sum_{t\in\left[T\right]}p\left(y=i\mid x;\theta^t\right)$ is the average predicted output over $T$ times of applying dropout to the model. $\theta^t$ denotes the parameters of the $t$-th dropout. We set $T=10$ with probability 0.1 at test time to sample different DNNs as \cite{malanie2018dfal}.

\textbf{5) LC} Least confidence only considers the most probable label of an instance to compute the uncertainty. That is, it ranks data based on the highest posterior probability and select data with the least confidence on the prediction:
\begin{equation}
    \label{equ_lc}
    \underset{{\bf{x}}\in{\mathcal{U}}}{\argmax}\left(1-p\left(y=y'\mid x;\theta\right)\right)
\end{equation}
where $y'=\underset{{\bf{i}}\in\left[C\right]}{\argmax}\left(p\left(y=i\mid x;\theta\right)\right)$ indicating the predicted class label by $f_\theta$.

\textbf{6) Margin} Since LC only utilizes the most confident class label, the information about the remaining labels is discarded but can be useful as well. To solve this issue, the margin sampling \cite{Scheffer2001ActiveHM} ranks data based on the difference between the most confident and second most confident labels and chooses the data with a small difference: 
\begin{equation}
    \label{equ_margin}
    \underset{{\bf{x}}\in{\mathcal{U}}}{\argmin}\left(\underset{{\bf{i}}\in\left[C\right]}{\argmax}\left(p\left(y=i\mid x;\theta\right)\right)-\underset{{\bf{i}}\in\left[C\right]/y'}{\argmax}\left(p\left(y=i\mid x;\theta\right)\right)\right)
\end{equation}
The hypothesis is that if a DNN predicts a similar probability on the top two labels for an instance, then this instance is not well learned and remains close to the decision boundary.

\textbf{7) MCP$^*$} Noticing that the data selected by margin sampling might be unbalanced distributed concerning the decision boundary areas, the Multiple-Boundary Clustering and Prioritization (MCP) \cite{Shen2020mcp} improves the margin sampling by uniformly selecting data from different areas. Besides, it computes the priority by the ratio of probabilities of the top two labels instead of using difference. Each decision boundary area is a cluster and is defined by the top two classes. Data are selected from each cluster with high priorities.

\textbf{8) DFAL} To approximate how close an instance is to the decision boundary, the DeepFool active learning (DFAL) \cite{malanie2018dfal} utilizes the magnitude of the minimum perturbation to successfully craft an adversarial example of this instance by the DeepFool attack. Indeed, adversarial attacks are designed to push the clean data to cross the decision boundary. The object of DFAL is 
\begin{equation}
\label{equ_dfal}
\underset{{\bf{x}}\in{\mathcal{U}}}{\argmin}\left(DeepFool\left(x,f_\theta,L_p\right)\right)
\end{equation}
where $DeepFool\left(x,f_\theta,L_p\right)$ is a function that attacks $f_\theta$ given $x$ using the $L_p$ norm ($p=2$) and outputs the perturbation between $x$ and its adversarial example. For the experiments, first, we borrow the provided implementation based on Keras by \cite{malanie2018dfal} and modify it to fit in PyTorch. Second, as DFAL starts with a set of randomly selected data in each stage, the size of data is set as 10 times the number of new data to label as suggested in the provided implementation. Third, to ensure a fair comparison, we discard the generated adversarial examples by DeepFool.

\textbf{9) EGL} Given that the DNN training generally uses gradient-based optimizations, the expected gradient length (EGL) \cite{burr2008EGL} ranks an instance with high importance if it would induce the greatest change in the gradient. A challenge is that in active learning, the labels are not available, thus, EGL assumes all the labels to data and computes the expected gradient. The data are selected by
\begin{equation}
    \label{equ_egl}
    \underset{{\bf{x}}\in{\mathcal{U}}}{\argmax}\left(\sum_{i\in\left[C\right]}p\left(y=i\mid x;\theta\right)\parallel\triangledown J\left(\theta,x,y=i\right)\parallel\right)
\end{equation}
where $\parallel\cdot\parallel$ is the $L_2$ norm (Euclidean distance), $\triangledown J\left(\theta,x,y=i\cdot\right)$ denotes the gradient of the loss function at $x$ given $\theta$, and $y=i$. 

Different from focusing on minimizing the maximum of the risk, some acquisition functions solve the optimization object in Equation \ref{equ_al} via adding data that are far from the labeled data, such as Core-set.

\textbf{10) Core-set} The Core-set selection \cite{sener2018active} sets an upper bound to the optimization object in Equation \ref{equ_al} and converts the problem to be equivalent to the k-center problem and is solved by selecting the data that are far from data in the labeled pool $\mathcal{L}$. Given that the selection of data is time-consuming, we apply the k-Center-Greedy instead of robust k-Center since they exhibit similar behavior \cite{sener2018active}.

\textbf{11) Random} Random sampling is the simplest and model-free method that each data has an equal probability of selection. This method, strictly speaking, belongs to passive learning but is commonly taken as a baseline in active learning. Due to the inherent randomness, we repeat our experiments three times but observed negligible variance in the results.

Note that the functions with $*$ are originally designed for test selection, but they all select the most informative data. Thus, they are suitable for active learning, and we include them in our study.

\section{Effectiveness of Robust Active Learning}
We investigate the effectiveness of robust active learning in creating models that are both accurate and robust. We use the acquisition functions and the experimental protocol that were described above. 

\begin{figure*}[htpb]
    \includegraphics[scale=0.3]{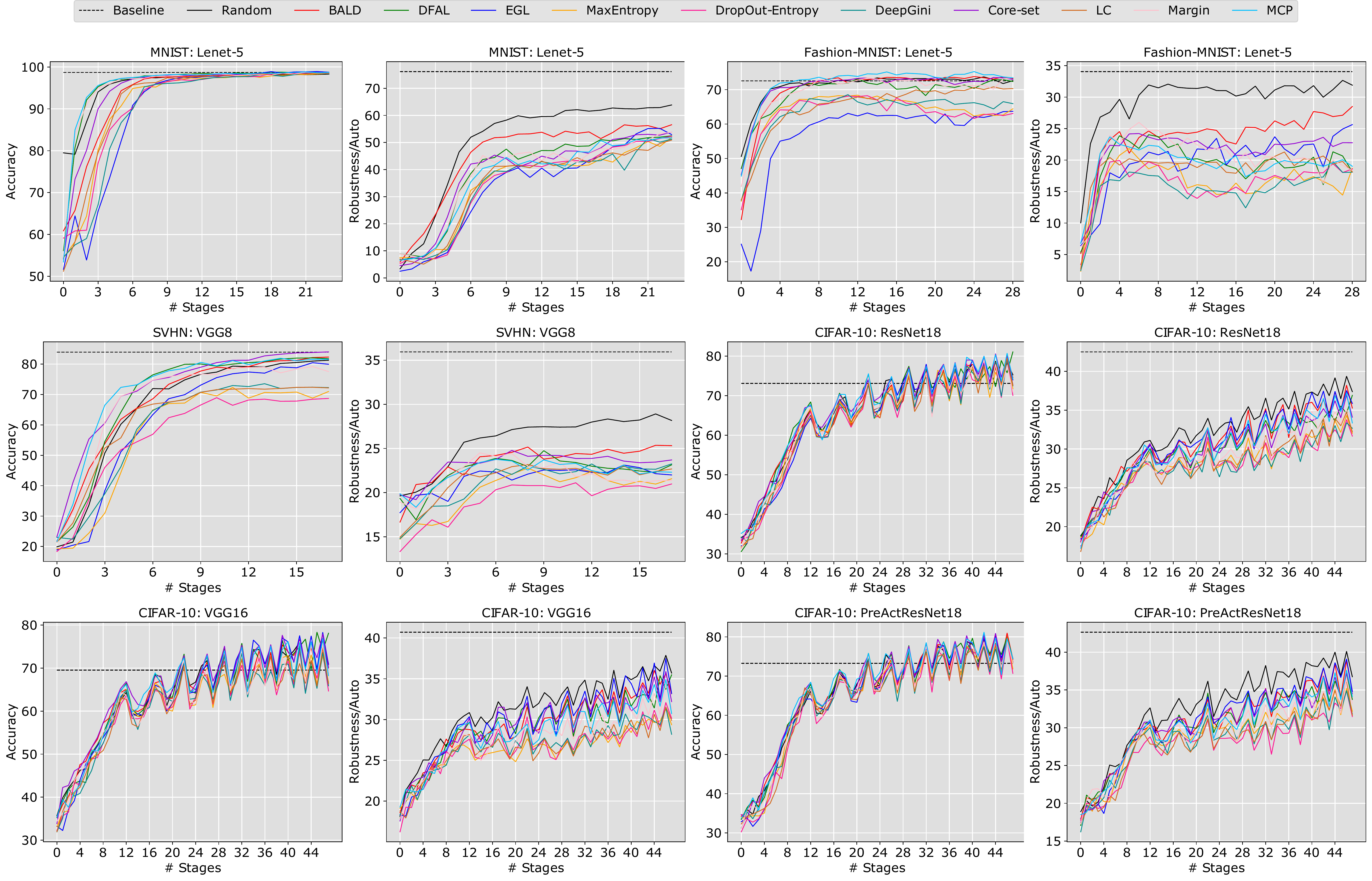}%
    \caption{Test accuracy (1st and 3rd columns) and robustness against Auto attack (2nd and 4th columns) of 11 acquisition functions over different stages of robust active learning. Baseline is the model adversarially trained using the entire data.}
    \label{fig:exp_2}
\end{figure*}

\emph{\textbf{Results:}} Figure \ref{fig:exp_2} visualizes the accuracy and robustness curves of 11 acquisition functions, for each dataset and model architecture we considered. First, we observe that most of the acquisition functions keep the ability to achieve the same level of test accuracy as the model adversarially trained with the full dataset. In general, 3\%, 2\%, 14\%, and 22\% of the labeled data are enough to reach this level of accuracy for MNIST, Fashion-MNIST, SVHN, and CIFAR-10, respectively. Some functions (e.g. MaxEntropy, Entropy, DeepGini, LC, and EGL), however, perform worse than the others -- including random sampling. MCP, Core-set, BALD, and (occasionally) DFAL are the best performing metrics and outperform random sampling.

Second, all acquisition functions allow a substantial increase of robustness compared to models trained with original data only. However, the obtained models are less robust than the model adversarially trained with all data (between 2\% and 10\% robustness difference compared to the best performing acquisition function). Interestingly, we observe that random sampling stands out in all the datasets and DNNs, and often by a significant margin. Just like for accuracy, MaxEntropy, Entropy, DeepGini, LC, and EGL, achieve the lowest levels of robustness. The other acquisition functions perform irregularly across different datasets and DNNs, yet remain inferior to random sampling. This indicates that the criteria used by existing acquisition functions have a negative effect on robustness, to the extent that random sampling -- a simple, random, model-free, and data-free method -- performs much better. 

To confirm this conclusion, we conduct statistical analysis and assess whether the difference of robustness between random sampling and the other acquisition functions is statistically significant. The analysis is based on the Wilcoxon signed-rank test \cite{wilcoxon1945}, which is a non-parametric statistical hypothesis test commonly used for comparing two independent paired samples. In our case, each sample is the robustness obtained based on a given acquisition function. Random sampling is compared with each of the other 10 functions on all datasets/models and three attacks. This yields 180 ($10 \times 6 \times 3$) statistical tests. We set the significance level $\alpha$ to $5.00E-02$. All these tests have rejected the null hypothesis that there is no difference between random sampling and the other acquisition function, with a p-value ranging from 8.15E-10 to 9.49E-03. Hence, we conclude that random sampling significantly outperforms the other functions in terms of robustness.

\vspace{2mm}
\noindent\fcolorbox{black}{answercolor}{\begin{minipage}{0.97\columnwidth}
\textbf{Conclusions:} Using only a limited set of labeled data, robust active learning can produce models with the same level of accuracy as the model adversarially trained with all data. When it comes to robustness, random sampling performs consistently better than the other acquisition functions, though there remains a substantial margin for improvement compared to the robustness of the model trained using all data.
\end{minipage}}

\section{Data Selection Biases}
As an attempt to explain the better effectiveness of random sampling and devise an improved acquisition function, we investigate the characteristics of the data selected by each acquisition function at all stages of the active learning process. We hypothesize that random sampling performs better than existing acquisition functions because the latter are \emph{biased} towards the ``most'' informative data -- informative being defined by each function in terms of the metric it uses to prioritize the data. By contrast, random sampling is naturally unbiased as it associates each data with the same probability to be selected. Therefore, on average it selects a set of data that are representative of the whole unlabeled pool.

We aim at establishing a relationship between the biases that the acquisition functions introduce and their inability to produce robust models. We study four data characteristics, entropy, Gini impurity, least confidence (lc), and margin, from the corresponding acquisition functions. The characteristics of the others are not applicable due to the use of dropouts (BALD, DropOut-Entropy) and model independence (MCP, DFAL, EGL, Core-set). Additionally, we measure the bias in the true class label, which is available at first hand and has been studied in the literature \cite{Prabhu2019bias}.

First, we measure the bias of each acquisition function towards each of these characteristics. Given a dataset and DNN, let $\mathcal{U}_{i,j}$ be the unlabeled pool in the $i$-th stage of active learning using the $j$-th acquisition function and $\mathcal{S}_{i,j} \subseteq \mathcal{U}_{i,j}$ be the set of data selected by $j$ at this $i$-th stage. Given a characteristic function $\phi$ that, given an input data $x$ returns the value of the characteristic under study for $x$, we generate two sets of variables $\phi\left(\mathcal{S}_{i,j}\right)$ and $\phi\left(\mathcal{U}_{i,j}\right)$ that contain the characteristic value of all data in $\mathcal{S}_{i,j}$ and $\mathcal{U}_{i,j}$, respectively. We, then, estimate the probability density functions (PDFs) of $\phi\left(\mathcal{S}_{i,j}\right)$ and $\phi\left(\mathcal{U}_{i,j}\right)$ using the histogram method with 50 bins \cite{silverman1986book}, yielding $PDF_{\phi\left(\mathcal{S}_{i,j}\right)}$ and $PDF_{\phi\left(\mathcal{U}_{i,j}\right)}$, respectively. We calculate the difference between two PDFs based on the Jensen-Shannon divergence (JSD) -- an established method to measure the divergence between two distributions \cite{Lin1991jsd}. This difference is, therefore, given by
\begin{equation}
    \label{equ:jsd}
    d_{i,j}=\mathrm{JSD}\left(PDF_{\phi\left(\mathcal{S}_{i,j}\right)}\parallel PDF_{\phi\left(\mathcal{U}_{i,j}\right)}\right)
\end{equation}

Second, we measure the correlation at each stage for every characteristic between (a) the robustness of models produced by different acquisition functions and (b) the characteristic bias of data selected by corresponding functions. We separate the different stages because, in the early stages, the learned model weights are not yet stable and may disrupt the results. To measure this correlation we use the Pearson correlation, which captures linear relationships between two variables. Hence, for each stage $i$ and each characteristic, we obtain a correlation coefficient $r_i$ defined as: \cite{student1908pearson}:
\begin{equation}
    \label{equ:pearson}
    r_i=\mathrm{Pearson}\left(d_i, a_i\right), d_i=\left\{d_{i,j}\right\}_{j\leq 11}, a_i=\left\{a_{i,j}\right\}_{j\leq 11}
\end{equation}
where $a_{i,j}$ denotes the robustness of the model produced at the $i$-th stage by the $j$-th acquisition function.

\begin{figure}
    \centering
    \includegraphics[scale=0.45]{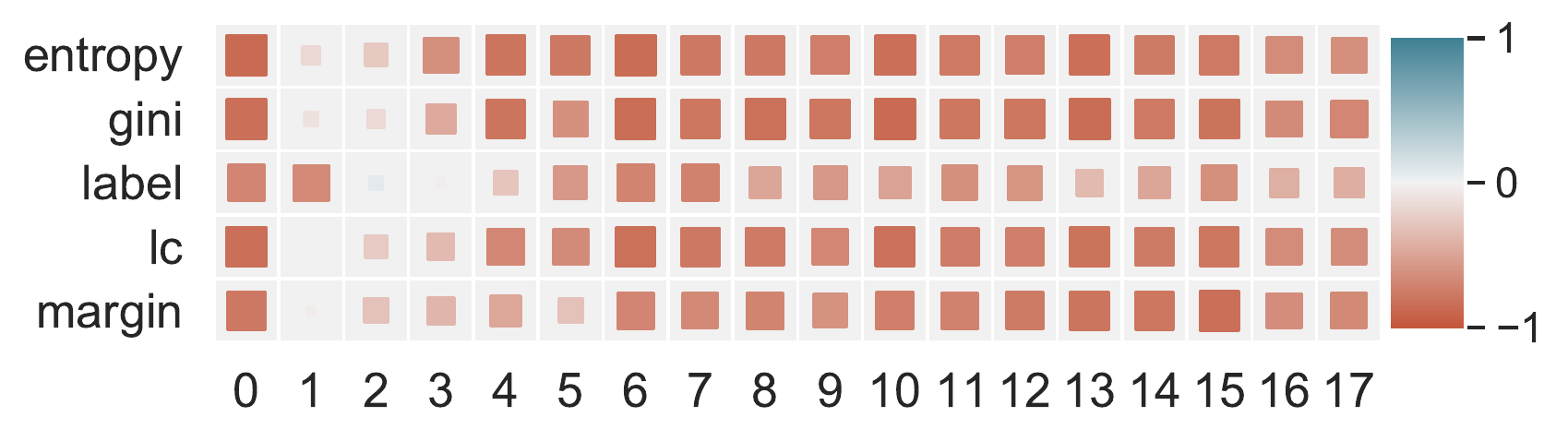}
    \caption{Heat map of Pearson correlation between bias of characteristics adversarial robustness against the Auto attack. $x$-axis: stages in active learning; $y$-axis: characteristics. The size of each square corresponds to the magnitude of the correlation it represents. Dataset: SVHN; DNN: VGG8.}
    \label{fig:char}
\end{figure}

\textbf{Results:} Figure \ref{fig:char} shows a heat map of the Pearson correlation coefficients $\left\{r_i\right\}$ obtained from SVHN/VGG8, and the Auto attack. We observe that all correlations are negative, which supports our hypothesis that characteristic biases have a negative effect on robustness. In other words, selecting data that are more representative of the entire unlabeled pool is more likely to produce a robust DNN. Among all the characteristics, entropy exhibits the strongest negative correlations. Over the 18 stages, the maximum and average correlations obtained by entropy are -0.84 and -0.65 (strong negative correlations), whereas the lowest correlations are achieved by margin (maximum -0.76 and -0.56 on average). Entropy, therefore, seems the most capable metric to drive the selection of a representative set of data that will be used to produce robust models. 

\noindent\fcolorbox{black}{answercolor}{\begin{minipage}{0.97\columnwidth}
\textbf{Conclusions:} The inherent bias of acquisition functions towards what they define as the ``most informative'' data has strong negative correlations with robustness. To improve robustness, an ideal acquisition function should rather select a set of data that have a representative level of informativeness. Among the informative characteristics we investigated, entropy bias has the strongest negative correlations with robustness.
\end{minipage}}

\section{Density-based Robust Sampling with Entropy}

\subsection{Definition and Algorithm}
Inspired by our previous results, we propose a new acquisition function for robust active learning: the density-based robust sampling with entropy (DRE). The key principle of DRE is to maintain a balance between selected data and unlabeled pool in terms of entropy distribution (represented by entropy PDF as defined in the previous section).

Algorithm \ref{algo} formalizes our DRE procedure. Given the initial labeled pool $\mathcal{L}$, the DNN parameters $\theta$ are obtained by training on $\mathcal{L}$ (Line 1). Afterward, in each stage of robust active learning, DRE iterates to select the successive sets of data to label (Lines 2-9). Concretely, we compute the entropy score of each data in the unlabeled pool $x\in\mathcal{U}$ (Line 3). Next, we calculate the PDF of the entropy values $\mathcal{A}_k$, noted $PDF_{\mathcal{A}_{k}}$, via the histogram method with 50 bins \cite{silverman1986book} (Line 4). Data are then randomly selected from each density interval and added to $\mathcal{L}$ (Lines 5-7). Finally, $\theta$ is updated by adversarial training on $\mathcal{L}$ (Line 8).

\begin{algorithm}
\caption{DRE: density-based robust sampling with entropy}
\label{algo}
\DontPrintSemicolon
  \KwInput{$\mathcal{U}$/$\mathcal{L}$: initial unlabeled/labeled pool}
  \KwInput{$b$: budget}
  \KwInput{$n$: number of labeling in each query}
  \KwOutput{$\theta$: DNN parameters}
  \tcc{initialization}
  Initialize $\theta$ by training on $\mathcal{L}, k=0$ \\
  \While{$k<\lceil\frac{b}{n}\rceil$}
  { 
    $\mathcal{A}_k=\left\{H\left(y,x;\theta\right), x\in\mathcal{U}\right\}$ \tcp*{compute entropy scores}
    $PDF_{\mathcal{A}_{k}}=DensityEstimation\left(\mathcal{A}_k\right)$ \tcp*{estimate PDF}
    \tcc{sample data from each density interval, $|\mathcal{S}_k|=n$}
    $\mathcal{S}_k=RandomSample\left(\mathcal{U}, PDF_{\mathcal{A}_{k}}\right)$ \\
    $\mathcal{U}=\mathcal{U}\backslash\mathcal{S}_k$ \tcp*{update unlabeled pool}
    $\mathcal{L}=\mathcal{L}\cup\mathcal{S}_k$ \tcp*{update labeled pool}
    Update $\theta$ by adversarial training on $\mathcal{L}$ \\
    $k=k+1$
  }
  Return $\theta$
\end{algorithm}

\subsection{Evaluation}
We evaluate DRE using the same datasets, DNNs, and experimental protocol that we previously used in Section 5.

\emph{\textbf{Results:}} Table \ref{tab:best} shows the accuracy and the robustness of the models produced by each acquisition function at one-third (1/3) and at the end (``last stage'') of the robust active learning process. We observe that DRE is competitive in terms of accuracy. At the end of the process, our acquisition function achieves the best accuracy for Fashion-MNIST and CIFAR-10/PreActResNet18, which is close to the best for MNIST and CIFAR-10/ResNet18, and ranks at about the middle for SVHN and CIFAR-10/VGG16. When it comes to robustness, DRE stands out and outperforms all the other acquisition functions for all models and datasets except CIFAR-10/VGG16 where it yields the robustness slightly lower than random sampling (-0.39\%). On the other datasets and models, DRE improves the robustness over random sampling by 0.75\% up to 3.84\%. 

\begin{table*}
\caption{Comparison of accuracy (\%) and robustness (\%) against the Auto attack between DRE and 11 acquisition functions. 1/3 stage: the 8th, 9th, 6th, 16th stages for MNIST, Fashion-MNIST, SVHN, and CIFAR-10, respectively. Last: the last stage. Highlight in gray: DRE is better than a function.}
\label{tab:best}
\resizebox{\textwidth}{!}{
\begin{tabular}{|l|rr|rr|rr|rr|rr|rr|}
\hline
 & \multicolumn{2}{c|}{\textbf{MNIST: Lenet-5}} & \multicolumn{2}{c|}{\textbf{Fashion-MNIST: Lenet-5}} & \multicolumn{2}{c|}{\textbf{SVHN: VGG8}} & \multicolumn{2}{c|}{\textbf{CIFAR-10: VGG16}} & \multicolumn{2}{c|}{\textbf{CIFAR-10: ResNet18}} & \multicolumn{2}{c|}{\textbf{CIFAR-10: PreActResNet18}} \\ \cline{2-13} 
 & \textbf{Accuracy} & \textbf{Robustness} & \textbf{Accuracy} & \textbf{Robustness} & \textbf{Accuracy} & \textbf{Robustness} & \textbf{Accuracy} & \textbf{Robustness} & \textbf{Accuracy} & \textbf{Robustness} & \textbf{Accuracy} & \textbf{Robustness} \\ \cline{2-13} 
\multirow{-3}{*}{\textbf{\begin{tabular}[c]{@{}l@{}}Acquisition\\ Function\end{tabular}}} & \multicolumn{12}{c|}{\textbf{1/3 stage}} \\ \hline
BALD & \cellcolor[HTML]{C0C0C0}97.15 & \cellcolor[HTML]{C0C0C0}50.13 & 73.09 & \cellcolor[HTML]{C0C0C0}24.15 & \cellcolor[HTML]{C0C0C0}65.51 & \cellcolor[HTML]{C0C0C0}24.06 & \cellcolor[HTML]{C0C0C0}60.68 & \cellcolor[HTML]{C0C0C0}28.17 & \cellcolor[HTML]{C0C0C0}60.61 & \cellcolor[HTML]{C0C0C0}29.02 & \cellcolor[HTML]{C0C0C0}64.79 & \cellcolor[HTML]{C0C0C0}29.29 \\
DFAL & 97.88 & \cellcolor[HTML]{C0C0C0}43.71 & \cellcolor[HTML]{C0C0C0}71.71 & \cellcolor[HTML]{C0C0C0}21.99 & 73.07 & \cellcolor[HTML]{C0C0C0}23.37 & \cellcolor[HTML]{C0C0C0}60.37 & \cellcolor[HTML]{C0C0C0}27.33 & \cellcolor[HTML]{C0C0C0}62.72 & \cellcolor[HTML]{C0C0C0}28.88 & \cellcolor[HTML]{C0C0C0}63.89 & \cellcolor[HTML]{C0C0C0}28.28 \\
EGL & \cellcolor[HTML]{C0C0C0}94.08 & \cellcolor[HTML]{C0C0C0}31.60 & \cellcolor[HTML]{C0C0C0}61.85 & \cellcolor[HTML]{C0C0C0}21.37 & \cellcolor[HTML]{C0C0C0}56.84 & \cellcolor[HTML]{C0C0C0}22.45 & \cellcolor[HTML]{C0C0C0}58.93 & \cellcolor[HTML]{C0C0C0}29.62 & \cellcolor[HTML]{C0C0C0}62.43 & \cellcolor[HTML]{C0C0C0}29.28 & \cellcolor[HTML]{C0C0C0}61.31 & \cellcolor[HTML]{C0C0C0}28.32 \\
MaxEntropy & \cellcolor[HTML]{C0C0C0}95.34 & \cellcolor[HTML]{C0C0C0}35.18 & \cellcolor[HTML]{C0C0C0}67.85 & \cellcolor[HTML]{C0C0C0}18.48 & \cellcolor[HTML]{C0C0C0}57.55 & \cellcolor[HTML]{C0C0C0}20.61 & \cellcolor[HTML]{C0C0C0}58.67 & \cellcolor[HTML]{C0C0C0}25.81 & \cellcolor[HTML]{C0C0C0}60.82 & \cellcolor[HTML]{C0C0C0}27.84 & \cellcolor[HTML]{C0C0C0}62.11 & \cellcolor[HTML]{C0C0C0}27.77 \\
DropOut-Entropy & \cellcolor[HTML]{C0C0C0}94.52 & \cellcolor[HTML]{C0C0C0}34.56 & \cellcolor[HTML]{C0C0C0}65.92 & \cellcolor[HTML]{C0C0C0}16.73 & \cellcolor[HTML]{C0C0C0}54.41 & \cellcolor[HTML]{C0C0C0}18.79 & \cellcolor[HTML]{C0C0C0}58.63 & \cellcolor[HTML]{C0C0C0}26.41 & \cellcolor[HTML]{C0C0C0}59.74 & \cellcolor[HTML]{C0C0C0}27.09 & \cellcolor[HTML]{C0C0C0}61.79 & \cellcolor[HTML]{C0C0C0}27.93 \\
DeepGini & \cellcolor[HTML]{C0C0C0}95.25 & \cellcolor[HTML]{C0C0C0}35.83 & \cellcolor[HTML]{C0C0C0}66.61 & \cellcolor[HTML]{C0C0C0}16.10 & \cellcolor[HTML]{C0C0C0}58.55 & \cellcolor[HTML]{C0C0C0}21.11 & \cellcolor[HTML]{C0C0C0}60.19 & \cellcolor[HTML]{C0C0C0}27.21 & \cellcolor[HTML]{C0C0C0}59.55 & \cellcolor[HTML]{C0C0C0}26.71 & \cellcolor[HTML]{C0C0C0}61.41 & \cellcolor[HTML]{C0C0C0}26.63 \\
Core-set & \cellcolor[HTML]{C0C0C0}97.46 & \cellcolor[HTML]{C0C0C0}42.77 & \cellcolor[HTML]{C0C0C0}72.33 & \cellcolor[HTML]{C0C0C0}23.51 & 71.10 & \cellcolor[HTML]{C0C0C0}23.34 & \cellcolor[HTML]{C0C0C0}61.21 & \cellcolor[HTML]{C0C0C0}26.99 & 64.30 & \cellcolor[HTML]{C0C0C0}28.24 & \cellcolor[HTML]{C0C0C0}63.59 & \cellcolor[HTML]{C0C0C0}29.49 \\
LC & \cellcolor[HTML]{C0C0C0}96.17 & \cellcolor[HTML]{C0C0C0}36.49 & \cellcolor[HTML]{C0C0C0}66.07 & \cellcolor[HTML]{C0C0C0}19.61 & \cellcolor[HTML]{C0C0C0}65.31 & \cellcolor[HTML]{C0C0C0}21.78 & \cellcolor[HTML]{C0C0C0}59.81 & \cellcolor[HTML]{C0C0C0}26.55 & \cellcolor[HTML]{C0C0C0}60.42 & \cellcolor[HTML]{C0C0C0}26.89 & \cellcolor[HTML]{C0C0C0}62.31 & \cellcolor[HTML]{C0C0C0}26.39 \\
Margin & \cellcolor[HTML]{C0C0C0}97.29 & \cellcolor[HTML]{C0C0C0}45.19 & 72.57 & \cellcolor[HTML]{C0C0C0}23.43 & 71.42 & \cellcolor[HTML]{C0C0C0}24.51 & \cellcolor[HTML]{C0C0C0}58.99 & \cellcolor[HTML]{C0C0C0}25.69 & 63.47 & \cellcolor[HTML]{C0C0C0}27.75 & \cellcolor[HTML]{C0C0C0}63.93 & \cellcolor[HTML]{C0C0C0}27.97 \\
MCP & 97.73 & \cellcolor[HTML]{C0C0C0}40.42 & 72.85 & \cellcolor[HTML]{C0C0C0}21.03 & 73.19 & \cellcolor[HTML]{C0C0C0}23.42 & 63.85 & \cellcolor[HTML]{C0C0C0}28.20 & \cellcolor[HTML]{C0C0C0}63.04 & \cellcolor[HTML]{C0C0C0}28.04 & \cellcolor[HTML]{C0C0C0}64.16 & \cellcolor[HTML]{C0C0C0}28.25 \\
Random & \cellcolor[HTML]{C0C0C0}97.51 & \cellcolor[HTML]{C0C0C0}54.10 & \cellcolor[HTML]{C0C0C0}71.84 & 32.07 & \cellcolor[HTML]{C0C0C0}65.48 & 26.19 & \cellcolor[HTML]{C0C0C0}59.24 & \cellcolor[HTML]{C0C0C0}29.47 & \cellcolor[HTML]{C0C0C0}63.33 & \cellcolor[HTML]{C0C0C0}30.32 & \cellcolor[HTML]{C0C0C0}63.90 & \cellcolor[HTML]{C0C0C0}31.01 \\
\textbf{DRE} & \textbf{97.61} & \textbf{56.00} & \textbf{72.51} & \textbf{31.75} & \textbf{68.27} & \textbf{25.95} & \textbf{61.35} & \textbf{30.83} & \textbf{63.43} & \textbf{31.38} & \textbf{65.17} & \textbf{31.76} \\ \hline
 & \multicolumn{12}{c|}{\textbf{Last stage}} \\ \hline
BALD & \cellcolor[HTML]{C0C0C0}98.56 & \cellcolor[HTML]{C0C0C0}56.55 & \cellcolor[HTML]{C0C0C0}73.19 & \cellcolor[HTML]{C0C0C0}28.51 & 82.31 & \cellcolor[HTML]{C0C0C0}25.32 & 71.02 & \cellcolor[HTML]{C0C0C0}33.17 & \cellcolor[HTML]{C0C0C0}74.55 & \cellcolor[HTML]{C0C0C0}35.27 & \cellcolor[HTML]{C0C0C0}74.28 & \cellcolor[HTML]{C0C0C0}34.83 \\
DFAL & \cellcolor[HTML]{C0C0C0}98.43 & \cellcolor[HTML]{C0C0C0}52.45 & \cellcolor[HTML]{C0C0C0}72.37 & \cellcolor[HTML]{C0C0C0}18.11 & 81.79 & \cellcolor[HTML]{C0C0C0}23.15 & 78.11 & 35.48 & 81.07 & \cellcolor[HTML]{C0C0C0}36.91 & \cellcolor[HTML]{C0C0C0}74.47 & \cellcolor[HTML]{C0C0C0}33.74 \\
EGL & 98.75 & \cellcolor[HTML]{C0C0C0}52.92 & \cellcolor[HTML]{C0C0C0}63.71 & \cellcolor[HTML]{C0C0C0}25.64 & \cellcolor[HTML]{C0C0C0}79.95 & \cellcolor[HTML]{C0C0C0}22.00 & 70.81 & \cellcolor[HTML]{C0C0C0}33.19 & \cellcolor[HTML]{C0C0C0}73.99 & \cellcolor[HTML]{C0C0C0}35.86 & \cellcolor[HTML]{C0C0C0}73.57 & \cellcolor[HTML]{C0C0C0}34.81 \\
MaxEntropy & \cellcolor[HTML]{C0C0C0}98.47 & \cellcolor[HTML]{C0C0C0}51.88 & \cellcolor[HTML]{C0C0C0}64.41 & \cellcolor[HTML]{C0C0C0}18.71 & \cellcolor[HTML]{C0C0C0}70.98 & \cellcolor[HTML]{C0C0C0}21.57 & \cellcolor[HTML]{C0C0C0}66.71 & \cellcolor[HTML]{C0C0C0}29.37 & \cellcolor[HTML]{C0C0C0}71.74 & \cellcolor[HTML]{C0C0C0}32.01 & \cellcolor[HTML]{C0C0C0}71.82 & \cellcolor[HTML]{C0C0C0}32.18 \\
DropOut-Entropy & 98.62 & \cellcolor[HTML]{C0C0C0}52.29 & \cellcolor[HTML]{C0C0C0}63.07 & \cellcolor[HTML]{C0C0C0}18.59 & \cellcolor[HTML]{C0C0C0}68.70 & \cellcolor[HTML]{C0C0C0}20.98 & \cellcolor[HTML]{C0C0C0}64.67 & \cellcolor[HTML]{C0C0C0}30.15 & \cellcolor[HTML]{C0C0C0}70.08 & \cellcolor[HTML]{C0C0C0}31.65 & \cellcolor[HTML]{C0C0C0}70.69 & \cellcolor[HTML]{C0C0C0}31.47 \\
DeepGini & \cellcolor[HTML]{C0C0C0}98.41 & \cellcolor[HTML]{C0C0C0}51.23 & \cellcolor[HTML]{C0C0C0}65.93 & \cellcolor[HTML]{C0C0C0}18.08 & \cellcolor[HTML]{C0C0C0}72.27 & \cellcolor[HTML]{C0C0C0}23.25 & \cellcolor[HTML]{C0C0C0}65.89 & \cellcolor[HTML]{C0C0C0}28.19 & \cellcolor[HTML]{C0C0C0}72.50 & \cellcolor[HTML]{C0C0C0}32.67 & \cellcolor[HTML]{C0C0C0}73.22 & \cellcolor[HTML]{C0C0C0}31.83 \\
Core-set & 98.63 & \cellcolor[HTML]{C0C0C0}53.82 & \cellcolor[HTML]{C0C0C0}72.99 & \cellcolor[HTML]{C0C0C0}22.76 & 84.01 & \cellcolor[HTML]{C0C0C0}23.70 & 70.09 & \cellcolor[HTML]{C0C0C0}32.20 & \cellcolor[HTML]{C0C0C0}72.74 & \cellcolor[HTML]{C0C0C0}34.09 & \cellcolor[HTML]{C0C0C0}74.53 & \cellcolor[HTML]{C0C0C0}34.49 \\
LC & \cellcolor[HTML]{C0C0C0}98.48 & \cellcolor[HTML]{C0C0C0}50.89 & \cellcolor[HTML]{C0C0C0}70.29 & \cellcolor[HTML]{C0C0C0}18.11 & \cellcolor[HTML]{C0C0C0}72.17 & \cellcolor[HTML]{C0C0C0}22.68 & \cellcolor[HTML]{C0C0C0}66.39 & \cellcolor[HTML]{C0C0C0}29.91 & \cellcolor[HTML]{C0C0C0}72.60 & \cellcolor[HTML]{C0C0C0}32.21 & \cellcolor[HTML]{C0C0C0}73.57 & \cellcolor[HTML]{C0C0C0}31.64 \\
Margin & 98.65 & \cellcolor[HTML]{C0C0C0}53.80 & \cellcolor[HTML]{C0C0C0}73.43 & \cellcolor[HTML]{C0C0C0}18.36 & \cellcolor[HTML]{C0C0C0}77.55 & \cellcolor[HTML]{C0C0C0}21.37 & \cellcolor[HTML]{C0C0C0}68.85 & \cellcolor[HTML]{C0C0C0}31.40 & \cellcolor[HTML]{C0C0C0}70.91 & \cellcolor[HTML]{C0C0C0}33.65 & \cellcolor[HTML]{C0C0C0}73.22 & \cellcolor[HTML]{C0C0C0}33.24 \\
MCP & 98.71 & \cellcolor[HTML]{C0C0C0}51.71 & \cellcolor[HTML]{C0C0C0}73.41 & \cellcolor[HTML]{C0C0C0}19.02 & 81.59 & \cellcolor[HTML]{C0C0C0}22.31 & \cellcolor[HTML]{C0C0C0}68.25 & \cellcolor[HTML]{C0C0C0}30.73 & \cellcolor[HTML]{C0C0C0}74.19 & \cellcolor[HTML]{C0C0C0}33.35 & \cellcolor[HTML]{C0C0C0}74.39 & \cellcolor[HTML]{C0C0C0}34.14 \\
Random & \cellcolor[HTML]{C0C0C0}98.30 & \cellcolor[HTML]{C0C0C0}63.85 & \cellcolor[HTML]{C0C0C0}72.61 & \cellcolor[HTML]{C0C0C0}31.90 & \cellcolor[HTML]{C0C0C0}81.28 & \cellcolor[HTML]{C0C0C0}28.17 & 70.03 & 35.70 & 75.29 & \cellcolor[HTML]{C0C0C0}37.38 & \cellcolor[HTML]{C0C0C0}73.28 & \cellcolor[HTML]{C0C0C0}36.75 \\
\textbf{DRE} & \textbf{98.59} & \textbf{67.69} & \textbf{73.79} & \textbf{33.79} & \textbf{81.41} & \textbf{29.81} & \textbf{69.35} & \textbf{35.31} & \textbf{74.99} & \textbf{38.66} & \textbf{74.69} & \textbf{37.86} \\ \hline
\end{tabular}
}
\end{table*}

We performed statistical tests to assess the statistical significance of the robustness improvement that DRE brings on all datasets, models, and learning stages. A Wilcoxon signed-rank test rejects the null hypothesis at the significance level of 5.00E-02 that there is no difference between DRE and any other acquisition function, with a p-value ranging from 8.15E-10 to 2.25E-03. 

\vspace{2mm}
\noindent\fcolorbox{black}{answercolor}{\begin{minipage}{0.97\columnwidth}
\textbf{Conclusions:} Our proposed DRE succeeds in consistently achieving better robustness than any other acquisition function -- and does so while remaining competitive in terms of accuracy. DRE, therefore, forms the baseline for robust active learning.
\end{minipage}}

\section{DRE for Test Selection}
While we previously showed that robust active learning can train robust models \emph{from scratch}, we study the adjacent problem of DL testing and retraining (DL T\&R). DL T\&R starts from a trained model and attempts to generate new test data that the model misclassifies. In turn, it uses these data to improve the model (through retraining) \cite{Wang2021robot}. Test generation methods generally proceed in two steps: 1) selection of the clean data to start from and 2) the generation of the test data from the clean data. For the second step, one can rely on dedicated methods that research has proposed \cite{deepxplore2017} or simply on adversarial attacks as \cite{yang2020gini}.

The acquisition functions used in active learning can also serve for test data selection (the first step mentioned above) and be combined with adversarial attacks algorithms to generate the data for retraining. Hence, we investigate the capability of these functions -- in particular, DRE -- to select data in a way that the robustness of the model will improve after retraining.

We utilize the same datasets and DNNs as previously. We pre-train the DNNs using standard training on the entire training set. Then, following established experimental protocols \cite{Wang2021robot}, we use a given acquisition function to select a budget test data (1\% to 10\% of the test set). We, then, generate adversarial examples from the selected data and retrain the DNNs using both the training data and the generated adversarial examples. Finally, we evaluate both the test accuracy and the adversarial robustness of the retrained DNNs. 

\emph{\textbf{Results:}} Table \ref{tab:best} presents the accuracy and robustness of each model, before and after retraining, against the Auto attack. The first conclusion we draw is that all acquisition functions are applicable as metrics in test selection. Compared with the baseline where no test data are added for training, they all manage to improve the robustness against the Auto attack by up to 3.23\% for SVHN and the three CIFAR-10 models. However, for MNIST and Fashion-MNIST, only DRE can increase the robustness of retrained DNNs by up to 2.58\%. Under different sizes of budget, DRE achieves competitive accuracy (the difference is less than 2.03\%) compared with the other 11 acquisition functions. Besides, concerning the robustness, DRE outperforms the others in most (116 of 144) cases by up to 8.21\%. 

\begin{table*}[]
\caption{Comparison of accuracy (\%)and robustness (\%) against the Auto attack with different budgets (1\%, 4\%) in test selection. Baseline: adversarially trained DNN using all training data. Highlight in gray: DRE is better than a function.}
\label{tab:test}
\resizebox{\textwidth}{!}{
\begin{tabular}{|l|rr|rr|rr|rr|rr|rr|rr|rr|rr|rr|rr|rr|}
\hline
 & \multicolumn{4}{c|}{\textbf{MNIST: Lenet-5}} & \multicolumn{4}{c|}{\textbf{Fashion-MNIST: Lenet-5}} & \multicolumn{4}{c|}{\textbf{SVHN: VGG8}} & \multicolumn{4}{c|}{\textbf{CIFAR-10: VGG16}} & \multicolumn{4}{c|}{\textbf{CIFAR-10: ResNet18}} & \multicolumn{4}{c|}{\textbf{CIFAR-10: PreActResNet18}} \\ \cline{2-25} 
 & \multicolumn{2}{r|}{\textbf{Accuracy}} & \multicolumn{2}{r|}{\textbf{Robustness}} & \multicolumn{2}{r|}{\textbf{Accuracy}} & \multicolumn{2}{r|}{\textbf{Robustness}} & \multicolumn{2}{r|}{\textbf{Accuracy}} & \multicolumn{2}{r|}{\textbf{Robustness}} & \multicolumn{2}{r|}{\textbf{Accuracy}} & \multicolumn{2}{r|}{\textbf{Robustness}} & \multicolumn{2}{r|}{\textbf{Accuracy}} & \multicolumn{2}{r|}{\textbf{Robustness}} & \multicolumn{2}{r|}{\textbf{Accuracy}} & \multicolumn{2}{r|}{\textbf{Robustness}} \\ \cline{2-25} 
\multirow{-3}{*}{\textbf{\begin{tabular}[c]{@{}l@{}}Acquisition\\ Function\end{tabular}}} & \textbf{1\%} & \textbf{4\%} & \textbf{1\%} & \textbf{4\%} & \textbf{1\%} & \textbf{4\%} & \textbf{1\%} & \textbf{4\%} & \textbf{1\%} & \textbf{4\%} & \textbf{1\%} & \textbf{4\%} & \textbf{1\%} & \textbf{4\%} & \textbf{1\%} & \textbf{4\%} & \textbf{1\%} & \textbf{4\%} & \textbf{1\%} & \textbf{4\%} & \textbf{1\%} & \textbf{4\%} & \textbf{1\%} & \textbf{4\%} \\ \hline
Baseline & \multicolumn{2}{c|}{\textit{98.69}} & \multicolumn{2}{c|}{\textit{76.17}} & \multicolumn{2}{c|}{\textit{72.55}} & \multicolumn{2}{c|}{\textit{34.06}} & \multicolumn{2}{c|}{\textit{83.95}} & \multicolumn{2}{c|}{\textit{35.94}} & \multicolumn{2}{c|}{\textit{69.54}} & \multicolumn{2}{c|}{\textit{40.71}} & \multicolumn{2}{c|}{\textit{73.11}} & \multicolumn{2}{c|}{\textit{42.65}} & \multicolumn{2}{c|}{\textit{73.24}} & \multicolumn{2}{c|}{\textit{42.65}} \\
BALD & 98.38 & \cellcolor[HTML]{C0C0C0}98.42 & \cellcolor[HTML]{C0C0C0}74.80 & \cellcolor[HTML]{C0C0C0}75.17 & \cellcolor[HTML]{C0C0C0}66.79 & \cellcolor[HTML]{C0C0C0}66.02 & \cellcolor[HTML]{C0C0C0}32.12 & \cellcolor[HTML]{C0C0C0}33.77 & \cellcolor[HTML]{C0C0C0}86.23 & 87.49 & \cellcolor[HTML]{C0C0C0}37.25 & \cellcolor[HTML]{C0C0C0}37.69 & \cellcolor[HTML]{C0C0C0}76.04 & 76.01 & \cellcolor[HTML]{C0C0C0}43.11 & \cellcolor[HTML]{C0C0C0}43.39 & \cellcolor[HTML]{C0C0C0}79.97 & \cellcolor[HTML]{C0C0C0}79.92 & \cellcolor[HTML]{C0C0C0}45.42 & 45.49 & \cellcolor[HTML]{C0C0C0}79.97 & 78.91 & \cellcolor[HTML]{C0C0C0}45.42 & 45.49 \\
DFAL & \cellcolor[HTML]{C0C0C0}98.23 & \cellcolor[HTML]{C0C0C0}98.53 & \cellcolor[HTML]{C0C0C0}74.22 & \cellcolor[HTML]{C0C0C0}75.73 & \cellcolor[HTML]{C0C0C0}65.64 & \cellcolor[HTML]{C0C0C0}65.53 & \cellcolor[HTML]{C0C0C0}32.28 & \cellcolor[HTML]{C0C0C0}33.61 & \cellcolor[HTML]{C0C0C0}86.34 & 86.64 & \cellcolor[HTML]{C0C0C0}36.61 & \cellcolor[HTML]{C0C0C0}35.78 & \cellcolor[HTML]{C0C0C0}75.27 & \cellcolor[HTML]{C0C0C0}75.73 & \cellcolor[HTML]{C0C0C0}43.01 & \cellcolor[HTML]{C0C0C0}42.88 & \cellcolor[HTML]{C0C0C0}80.39 & \cellcolor[HTML]{C0C0C0}80.31 & \cellcolor[HTML]{C0C0C0}44.86 & \cellcolor[HTML]{C0C0C0}45.01 & \cellcolor[HTML]{C0C0C0}80.39 & 78.89 & \cellcolor[HTML]{C0C0C0}44.86 & \cellcolor[HTML]{C0C0C0}45.01 \\
EGL & \cellcolor[HTML]{C0C0C0}98.26 & \cellcolor[HTML]{C0C0C0}98.30 & \cellcolor[HTML]{C0C0C0}73.96 & \cellcolor[HTML]{C0C0C0}73.60 & \cellcolor[HTML]{C0C0C0}64.93 & \cellcolor[HTML]{C0C0C0}65.16 & \cellcolor[HTML]{C0C0C0}30.54 & \cellcolor[HTML]{C0C0C0}31.91 & \cellcolor[HTML]{C0C0C0}86.13 & 88.43 & \cellcolor[HTML]{C0C0C0}35.51 & \cellcolor[HTML]{C0C0C0}35.12 & \cellcolor[HTML]{C0C0C0}75.77 & 76.47 & \cellcolor[HTML]{C0C0C0}42.81 & \cellcolor[HTML]{C0C0C0}42.60 & \cellcolor[HTML]{C0C0C0}80.37 & \cellcolor[HTML]{C0C0C0}79.09 & \cellcolor[HTML]{C0C0C0}44.53 & \cellcolor[HTML]{C0C0C0}44.49 & \cellcolor[HTML]{C0C0C0}80.37 & 79.12 & \cellcolor[HTML]{C0C0C0}44.53 & \cellcolor[HTML]{C0C0C0}44.49 \\
MaxEntropy & \cellcolor[HTML]{C0C0C0}98.25 & \cellcolor[HTML]{C0C0C0}98.42 & \cellcolor[HTML]{C0C0C0}74.99 & \cellcolor[HTML]{C0C0C0}75.68 & \cellcolor[HTML]{C0C0C0}63.77 & \cellcolor[HTML]{C0C0C0}64.07 & \cellcolor[HTML]{C0C0C0}32.11 & \cellcolor[HTML]{C0C0C0}30.76 & \cellcolor[HTML]{C0C0C0}86.35 & 87.42 & \cellcolor[HTML]{C0C0C0}36.34 & \cellcolor[HTML]{C0C0C0}35.96 & \cellcolor[HTML]{C0C0C0}76.14 & 76.27 & \cellcolor[HTML]{C0C0C0}42.67 & \cellcolor[HTML]{C0C0C0}42.87 & \cellcolor[HTML]{C0C0C0}80.53 & \cellcolor[HTML]{C0C0C0}80.29 & \cellcolor[HTML]{C0C0C0}45.09 & \cellcolor[HTML]{C0C0C0}44.79 & \cellcolor[HTML]{C0C0C0}80.53 & 79.01 & \cellcolor[HTML]{C0C0C0}45.09 & \cellcolor[HTML]{C0C0C0}44.79 \\
DropOut-Entropy & 98.39 & \cellcolor[HTML]{C0C0C0}98.40 & \cellcolor[HTML]{C0C0C0}75.49 & \cellcolor[HTML]{C0C0C0}75.37 & \cellcolor[HTML]{C0C0C0}65.33 & \cellcolor[HTML]{C0C0C0}66.19 & \cellcolor[HTML]{C0C0C0}31.17 & \cellcolor[HTML]{C0C0C0}32.05 & \cellcolor[HTML]{C0C0C0}85.64 & 87.69 & \cellcolor[HTML]{C0C0C0}36.65 & \cellcolor[HTML]{C0C0C0}36.36 & \cellcolor[HTML]{C0C0C0}75.69 & 76.03 & \cellcolor[HTML]{C0C0C0}43.28 & \cellcolor[HTML]{C0C0C0}42.87 & \cellcolor[HTML]{C0C0C0}80.15 & \cellcolor[HTML]{C0C0C0}80.17 & \cellcolor[HTML]{C0C0C0}45.13 & \cellcolor[HTML]{C0C0C0}45.11 & \cellcolor[HTML]{C0C0C0}80.15 & 79.29 & \cellcolor[HTML]{C0C0C0}45.13 & \cellcolor[HTML]{C0C0C0}45.11 \\
DeepGini & 98.32 & \cellcolor[HTML]{C0C0C0}98.19 & \cellcolor[HTML]{C0C0C0}74.65 & \cellcolor[HTML]{C0C0C0}74.43 & \cellcolor[HTML]{C0C0C0}67.48 & \cellcolor[HTML]{C0C0C0}63.65 & \cellcolor[HTML]{C0C0C0}31.37 & \cellcolor[HTML]{C0C0C0}33.43 & 86.71 & 87.01 & \cellcolor[HTML]{C0C0C0}35.60 & \cellcolor[HTML]{C0C0C0}36.14 & \cellcolor[HTML]{C0C0C0}76.09 & 75.77 & \cellcolor[HTML]{C0C0C0}42.97 & \cellcolor[HTML]{C0C0C0}43.07 & \cellcolor[HTML]{C0C0C0}80.32 & \cellcolor[HTML]{C0C0C0}80.09 & \cellcolor[HTML]{C0C0C0}45.04 & \cellcolor[HTML]{C0C0C0}44.78 & \cellcolor[HTML]{C0C0C0}80.32 & 79.01 & \cellcolor[HTML]{C0C0C0}45.04 & \cellcolor[HTML]{C0C0C0}44.78 \\
Core-set & \cellcolor[HTML]{C0C0C0}98.31 & \cellcolor[HTML]{C0C0C0}98.13 & \cellcolor[HTML]{C0C0C0}75.27 & \cellcolor[HTML]{C0C0C0}74.59 & \cellcolor[HTML]{C0C0C0}64.53 & \cellcolor[HTML]{C0C0C0}65.77 & \cellcolor[HTML]{C0C0C0}30.53 & \cellcolor[HTML]{C0C0C0}31.51 & \cellcolor[HTML]{C0C0C0}85.59 & 86.65 & \cellcolor[HTML]{C0C0C0}36.83 & \cellcolor[HTML]{C0C0C0}37.15 & \cellcolor[HTML]{C0C0C0}75.71 & \cellcolor[HTML]{C0C0C0}75.70 & \cellcolor[HTML]{C0C0C0}43.19 & \cellcolor[HTML]{C0C0C0}43.51 & \cellcolor[HTML]{C0C0C0}78.83 & \cellcolor[HTML]{C0C0C0}78.94 & \cellcolor[HTML]{C0C0C0}45.53 & 45.58 & \cellcolor[HTML]{C0C0C0}78.83 & 78.75 & \cellcolor[HTML]{C0C0C0}45.53 & 45.58 \\
LC & \cellcolor[HTML]{C0C0C0}98.21 & \cellcolor[HTML]{C0C0C0}98.33 & \cellcolor[HTML]{C0C0C0}74.03 & \cellcolor[HTML]{C0C0C0}75.29 & \cellcolor[HTML]{C0C0C0}64.86 & \cellcolor[HTML]{C0C0C0}65.71 & \cellcolor[HTML]{C0C0C0}31.83 & \cellcolor[HTML]{C0C0C0}32.03 & 86.71 & 87.67 & \cellcolor[HTML]{C0C0C0}35.78 & \cellcolor[HTML]{C0C0C0}35.99 & \cellcolor[HTML]{C0C0C0}75.64 & \cellcolor[HTML]{C0C0C0}75.57 & \cellcolor[HTML]{C0C0C0}42.91 & \cellcolor[HTML]{C0C0C0}42.89 & \cellcolor[HTML]{C0C0C0}79.97 & \cellcolor[HTML]{C0C0C0}79.99 & \cellcolor[HTML]{C0C0C0}45.17 & \cellcolor[HTML]{C0C0C0}45.00 & \cellcolor[HTML]{C0C0C0}79.97 & 78.85 & \cellcolor[HTML]{C0C0C0}45.17 & \cellcolor[HTML]{C0C0C0}45.00 \\
Margin & \cellcolor[HTML]{C0C0C0}98.31 & \cellcolor[HTML]{C0C0C0}98.40 & \cellcolor[HTML]{C0C0C0}75.07 & \cellcolor[HTML]{C0C0C0}75.15 & \cellcolor[HTML]{C0C0C0}65.27 & \cellcolor[HTML]{C0C0C0}65.19 & \cellcolor[HTML]{C0C0C0}32.17 & \cellcolor[HTML]{C0C0C0}31.35 & 86.71 & 86.45 & \cellcolor[HTML]{C0C0C0}36.27 & \cellcolor[HTML]{C0C0C0}36.38 & \cellcolor[HTML]{C0C0C0}75.74 & 75.84 & \cellcolor[HTML]{C0C0C0}42.87 & \cellcolor[HTML]{C0C0C0}43.19 & \cellcolor[HTML]{C0C0C0}80.06 & \cellcolor[HTML]{C0C0C0}78.68 & \cellcolor[HTML]{C0C0C0}45.01 & \cellcolor[HTML]{C0C0C0}44.98 & \cellcolor[HTML]{C0C0C0}80.06 & 78.89 & \cellcolor[HTML]{C0C0C0}45.01 & \cellcolor[HTML]{C0C0C0}44.98 \\
MCP & \cellcolor[HTML]{C0C0C0}98.35 & \cellcolor[HTML]{C0C0C0}98.50 & \cellcolor[HTML]{C0C0C0}75.35 & \cellcolor[HTML]{C0C0C0}75.79 & \cellcolor[HTML]{C0C0C0}67.71 & \cellcolor[HTML]{C0C0C0}67.37 & \cellcolor[HTML]{C0C0C0}32.71 & \cellcolor[HTML]{C0C0C0}32.03 & 87.09 & 87.69 & \cellcolor[HTML]{C0C0C0}35.77 & \cellcolor[HTML]{C0C0C0}36.00 & \cellcolor[HTML]{C0C0C0}75.19 & \cellcolor[HTML]{C0C0C0}76.55 & \cellcolor[HTML]{C0C0C0}43.13 & \cellcolor[HTML]{C0C0C0}43.13 & \cellcolor[HTML]{C0C0C0}79.97 & \cellcolor[HTML]{C0C0C0}80.29 & \cellcolor[HTML]{C0C0C0}45.20 & \cellcolor[HTML]{C0C0C0}45.15 & \cellcolor[HTML]{C0C0C0}79.97 & 78.90 & \cellcolor[HTML]{C0C0C0}45.20 & \cellcolor[HTML]{C0C0C0}45.15 \\
Random & \cellcolor[HTML]{C0C0C0}98.21 & \cellcolor[HTML]{C0C0C0}98.18 & \cellcolor[HTML]{C0C0C0}74.28 & \cellcolor[HTML]{C0C0C0}75.08 & \cellcolor[HTML]{C0C0C0}65.72 & \cellcolor[HTML]{C0C0C0}68.14 & \cellcolor[HTML]{C0C0C0}29.55 & \cellcolor[HTML]{C0C0C0}28.43 & \cellcolor[HTML]{C0C0C0}85.37 & \cellcolor[HTML]{C0C0C0}85.93 & \cellcolor[HTML]{C0C0C0}37.29 & \cellcolor[HTML]{C0C0C0}37.09 & \cellcolor[HTML]{C0C0C0}75.65 & \cellcolor[HTML]{C0C0C0}75.61 & \cellcolor[HTML]{C0C0C0}43.45 & \cellcolor[HTML]{C0C0C0}43.30 & \cellcolor[HTML]{C0C0C0}80.29 & \cellcolor[HTML]{C0C0C0}80.03 & \cellcolor[HTML]{C0C0C0}45.03 & \cellcolor[HTML]{C0C0C0}45.29 & \cellcolor[HTML]{C0C0C0}80.29 & \cellcolor[HTML]{C0C0C0}78.69 & \cellcolor[HTML]{C0C0C0}45.03 & \cellcolor[HTML]{C0C0C0}45.29 \\
\textbf{DRE} & \textbf{98.32} & \textbf{98.54} & \textbf{76.52} & \textbf{76.88} & \textbf{69.46} & \textbf{69.64} & \textbf{34.56} & \textbf{36.64} & \textbf{86.58} & \textbf{86.40} & \textbf{37.36} & \textbf{38.46} & \textbf{76.34} & \textbf{75.76} & \textbf{43.62} & \textbf{43.94} & \textbf{82.54} & \textbf{82.60} & \textbf{45.58} & \textbf{45.40} & \textbf{82.54} & \textbf{78.72} & \textbf{45.58} & \textbf{45.40} \\ \hline
\end{tabular}
}
\end{table*}

\vspace{2mm}
\noindent\fcolorbox{black}{answercolor}{\begin{minipage}{0.97\columnwidth}
\textbf{Conclusions:} Although all acquisition functions are viable for DL testing and retraining, DRE yields higher robustness (by up to 8.21\%) than any of the other functions.\end{minipage}}

\section{Threats to Validity}
\label{sec:threat}
The internal threat of our work mainly comes from the implementation of compared acquisition functions and evaluation metrics. Regarding the 10 functions (except random sampling), we searched for the original implementations provided by the corresponding authors or the implementation by the other researchers, then carefully modified them into our framework based on PyTorch. For the involved parameters, we follow the suggestion in their implementations or publications. In the robustness evaluation and statistical analysis, we adopt popular libraries, such as Torchattacks and SciPy.

The external threats to validity are related to the selection of datasets, DNNs, and compared acquisition functions. For the datasets and DNNs, we employ four publicly available datasets and six DNN architectures that are widely studied in the experiments of different works about active learning. Regarding the compared acquisition functions, we include random sampling and ten well-designed functions in the literature, which cover the basic and recently proposed ones. Besides, these ten functions exist in both the machine learning and software engineering (e.g. MCP and DeepGini) communities.
 
The construct threats might be from the randomness and the robustness evaluation measures. To reduce the threat by randomness, each experiment is repeated three times and we present the average. To gauge the adversarial robustness, in the literature, we adopt two commonly used attacks (PGD and Auto attack, which are the current standard for robustness evaluation) and added the black-box gradient-free square attack. Though we show the results for the Auto attack only, the results for the other attacks (available on our companion website) corroborate our findings. Indeed, as the Auto attack is stronger than both the PGD attack and square attack, the robustness obtained by an acquisition function is lower when using the Auto attack to evaluate. Correspondingly, the relative difference will be slightly larger. For example, DRE achieves better robustness than random sampling by up to 6.89\% against the PGD attack (versus better by up to 3.84\% robustness against Auto attack). All detailed results are provided on our companion website \cite{homepage}.
 
\section{Conclusion}
We investigated the use of active learning for building robust deep learning systems. We conducted a comprehensive study with 11 existing acquisition functions and 15105 trained DNNs. We revealed that, in robust active learning, random sampling achieves better adversarial robustness than existing functions but fails to win on accuracy. Via our analyses, we demonstrated that the selected data for training a robust model should be representative of the entire candidate set and proposed DRE, the first acquisition function dedicated to robust active learning. Extensive experiments have demonstrated that DRE achieves better robustness than the compared acquisition functions -- including random -- and achieves competitive accuracy. Besides, the experiments on DL testing and retraining have demonstrated that DRE is suitable for this task and still outperforms the other acquisition functions. We hope that our formulation of the robust active learning process, the experimental protocol we put in place, and the baseline we propose (DRE) will altogether inspire future research on developing robust DL models. To support such research in uncovering more insights and developing even better approaches, our data and code\footnote{Code will be released upon acceptance.} are made publicly available.

\bibliographystyle{ACM-Reference-Format}

\clearpage
\bibliography{refs}

\end{document}